\documentclass[
]{ceurart}

\usepackage{listings}
\usepackage{verbatim}

\graphicspath{ {./figures/} }

\begin{document}

\copyrightyear{2023}
\copyrightclause{Copyright for this paper by its authors.
  Use permitted under Creative Commons License Attribution 4.0
  International (CC BY 4.0).}

\conference{Italian Workshop on Artificial Intelligence for Human-Machine Interaction (AIxHMI 2023), November 06, 2023, Rome, Italy}

\title{A multi-artifact EEG denoising by frequency-based deep learning}






\author[1,2]{Matteo Gabardi}[%
orcid=,
email=m.gabardi@campus.unimib.it,
url=,
]
\cormark[1]

\author[1,3]{Aurora Saibene}[%
orcid=0000-0002-4405-8234,
email=aurora.saibene@unimib.it,
url=https://mmsp.unimib.it/aurora-saibene/,
]

\author[1,3]{Francesca Gasparini}[%
orcid=0000-0002-6279-6660,
email=francesca.gasparini@unimib.it,
url=https://mmsp.unimib.it/francesca-gasparini/,
]

\author[2]{Daniele Rizzo}[%
orcid=,
email=drizzo@x-next.com,
url=,
]

\author[1,3]{Fabio Antonio Stella}[%
orcid=0000-0002-1394-0507,
email=fabio.stella@unimib.it,
url=,
]

\address[1]{University of Milano-Bicocca, Viale Sarca 336, 20126, Milano, Italy}
\address[2]{Xnext S.p.A., Via Valtorta 48, 20127, Milano, Italy}
\address[3]{NeuroMI, Milan Center for Neuroscience, Piazza dell’Ateneo Nuovo 1, 20126, Milano, Italy}

\cortext[1]{Corresponding author.}

\begin{abstract}
    Electroencephalographic (EEG) signals are fundamental to neuroscience research and clinical applications such as brain-computer interfaces and neurological disorder diagnosis. These signals are typically a combination of neurological activity and noise, originating from various sources, including physiological artifacts like ocular and muscular movements.
    Under this setting, we tackle the challenge of distinguishing neurological activity from noise-related sources.
    We develop a novel EEG denoising model that operates in the frequency domain, leveraging prior knowledge about noise spectral features to adaptively compute optimal convolutional filters for noise separation. 
    The model is trained to learn an empirical relationship connecting the spectral characteristics of noise and noisy signal to a non-linear transformation which allows signal denoising.
    Performance evaluation on the \emph{EEGdenoiseNet} dataset shows that the proposed model achieves optimal results according to both temporal and spectral metrics. The model is found to remove physiological artifacts from input EEG data, thus achieving effective EEG denoising.
    Indeed, the model performance either matches or outperforms that achieved by benchmark models, proving to effectively remove both muscle and ocular artifacts without the need to perform any training on the particular type of artifact.
  \end{abstract}


\maketitle


\section{Introduction}
The electroencephalographic (EEG) signal is a time series acquired with non-invasive sensors (called electrodes) placed on a subject's scalp and is characterized by time, frequency and spatial information \cite{saibene2023eeg}. However, it usually presents a mixture of neurological activity and signals deriving from noise-related biological or non-physiological sources \citep{zhang2020emotion}. \\
This means that besides recording neural signals, the EEG captures noise generated from ocular, muscular, and cardiac movements as examples of biological artifacts, and noise related to non-biological sources like cable movement, electrical interference, and electrode bad positioning \citep{uriguen2015eeg}. For further information on other artifact types, please refer to  the review papers by \textit{Urig{\"u}en and Garcia-Zapirain} \cite{uriguen2015eeg}, and \textit{Rashmi and Shantala} \cite{rashmi2022eeg}.  

Many attempts have been presented in the state-of-the-art to reduce or remove these types of artifacts, but automatic EEG denoising remains an open challenge \citep{jiang2019removal, mumtaz2021review}.
In particular, this paper focuses on the denoising of two specific types of artifacts, \textit{i.e.}, \emph{ocular} (OAs) and \emph{muscular artifacts} (MAs).

OAs are the most easily detectable artifacts due to their spiking shape similar to a \textit{V} and their pronounced presence on signals recorded by frontal electrodes \citep{wallstrom2004automatic}. Moreover, they have a frequency range between 0.5 and 3 Hz, and high amplitudes (around 100 mV) \citep{rashmi2022eeg}. Notice that reference electrodes, called electrooculograms (EOG), are sometimes included in the experimental setting to track eye movements. Similar references, \textit{i.e.}, electromyographic (EMG) sensors, can be also placed to detect surface muscular activity, which may introduce MAs \citep{uriguen2015eeg}. These artifacts are related to movements like swallowing, chewing, talking, clenching hands, and muscular tension \citep{rashmi2022eeg}. Unfortunately, MAs are more difficult to detect and present spectral characteristics overlapping with the neural ones, having that they usually have a frequency lesser than or equal to 35 Hz \citep{rashmi2022eeg}.\\
Notice that even if reference electrodes (EOG and EMG) can be used to track artifacts, the interference between them and the EEG related electrodes is bidirectional, i.e., the artifacts contaminate the EEG signals, and the EOG and EMG electrodes capture both artifacts and neural activity \cite{wallstrom2004automatic}. Thus the removal of OAs and MAs exploiting these sensors could be prone to errors or excessive neural signal removal \citep{wallstrom2004automatic}.
Therefore, the initial exploitation of EOG and EMG sensors in linear regression methods, shifted to methodologies like filtering, blind source separation, source decomposition, empirical mode decomposition, signal space projection, beamforming, and hybrid techniques \citep{uriguen2015eeg, mannan2018identification, chen2019removal}.
However, in the last few years, some works have proposed to move from the traditional denoising techniques previously reported to completely data-driven techniques based on deep learning (DL) models.
In this framework, this paper presents a DL-based denoising model, which relies on the knowledge related to the power spectral density (PSD) estimation of noisy EEG signals, and EOG or EMG noise data. In particular, the proposed model follows the guidelines provided by \textit{Zhang et al.} \cite{zhang2021eegdenoisenet}, who present \emph{EEGdenoiseNet}, i.e., a dataset devised as a benchmark to train and test DL-based denoising strategies. Therefore, the paper is structured as follows. \\
After a brief overview of current studies presenting DL-based denoising techniques (Section \ref{sec:related_works}), \emph{EEGdenoiseNet} is described to provide a better understanding of the exploited data (Section \ref{sec:dataset}). Afterwards, the proposed DL model is detailed (Section \ref{sec:proposed_model}). Section \ref{sec:results_discussion} reports the performed experiments, discuss both obtained and literature results, and provide future developments. Final considerations are presented in Section \ref{sec:conclusions}.

\section{Related works}\label{sec:related_works}
This study focuses on DL-based denoising techniques, which will be detailed in this section, and thus will not provide a dissertation of traditional processing methodologies. The readers are invited to consult insightful review papers on these topics provided by the EEG research community \cite{uriguen2015eeg, minguillon2017trends, mannan2018identification, chen2019removal}. 

Starting from the work related to the exploited dataset, \textit{Zhang et al.} \cite{zhang2021eegdenoisenet} do not produce only \emph{EEGdenoiseNet}, but also develop (i) a fully-connected neural network (FCNN), (ii) a simple CNN, (iii) a complex CNN, and (iv) a recurrent neural network (RNN) for benchmarking purposes. Moreover, in a later publication \citep{zhang2021novel}, the authors propose a novel CNN to remove MAs. In particular, the devised architecture is composed by seven blocks, of which the first six contain two 1D convolutional layers with ReLu as the activation function and a 1D average pooling layer. The last block has as well two 1D convolutional layers, which are instead followed by a flatten layer. Finally, a dense layer is inserted. Notice that the core of the proposal is related to the learning process. As reported by the authors, the aim of the DL-based denoising models is to define a function that projects the noisy signals to the clean ones:
\begin{equation}
    \tilde{x} = f(\hat{y}, \theta)
\end{equation}
where $\tilde{x}$ is the clean EEG, $\hat{y}$ is the normalized noisy EEG, and $\theta$ is the parameter to be learned.\\
Notice that the authors \cite{zhang2021eegdenoisenet} also report results obtained by applying traditional denoising techniques, i.e., empirical mode decomposition and filtering, and demonstrate that their DL-based model provides a better data denoising. Therefore, in this paper only comparisons with this benchmark model and other DL-based proposals will be provided.

Subsequently, other approaches have been proposed in the literature, like \textit{Yu et al.}'s end-to-end DL framework, called \emph{DeepSeparator} \cite{yu2022embedding}. This model is based on Inception-like blocks and composed of (i) an encoder deputed to feature extraction, (ii) a decomposer exploited to detect and remove OAs and MAs, and (iii) a decoder used to reconstruct the cleaned signal. \\
Notice that the authors propose a training strategy where three input and output pairs are designed to learn from both clean signals and artifacts: $\langle$ noisy EEG, clean EEG $\rangle$, $\langle$ clean EEG, clean EEG $\rangle$, and $\langle$ artifacts, artifacts $\rangle$.

Another proposal is \emph{EEGDnet} \cite{pu2022eegdnet}, which considers both non-local and local self-similarities of EEG signals. Notice that the model has a 2D transformer structure devised to remove OAs and MAs from 1D EEG signals. The clean and noise signals are summed up considering a specific signal-to-noise ratio (SNR) and the resulting noisy signals fed to EEGDnet. Afterwards, the input is reshaped in a 2D matrix and passed to a self-attention block, a normalization layer, a feed-forward block, and another normalization layer to finally reconstruct the signal.\\
Similarly, \textit{Wang, Li, and Wang} \cite{wang2022novel} propose a network mainly composed by a bidirectional gated recurrent unit, a self-attention, and a dense layer to remove OAs and MAs.

A Multi-Module Neural Network (MMNN) \cite{zhang2022novel} is developed to be used in real-time environments and considering single-channel EEG data. MMNN has a modular structure constituted by blocks containing convolutional and fully-connected layers. The model convergence and learning ability is supported by the residual connections intra- and inter-blocks.

Other proposals exploit Generative Adversarial Networks (GANs) to remove noise. For example, \textit{Brophy et al.} \cite{brophy2022denoising} sample the generator input directly from noisy EEG signals and make a comparison with the corresponding clean EEG signals in the discriminator. The generator is constituted by a Long-Short Term Memory (LSTM) network, while the discriminator is composed by four 1D convolutional layers and a fully-connected layer.\\
Similarly, \textit{Wang, Luo, and Shen} \cite{wang2022improved} generator consists of a Bidirectional-LSTM (BiLSTM) and a LSTM layer, while the discriminator comprises five CNN layers plus a fully-connected layer. The noisy EEG are passed to the generator, producing the denoised EEG, which is inputted to the discriminator with the ground truth data. Therefore, the authors' main aim is to map the relationships between clean EEG and artifacts to iteratively reduce the noise.

Finally, OAs only removal strategies are reported. \textit{Ozdemir, Kizilisik, and Guren} \cite{ozdemir2022removal} focus on the use of BiLSTM and propose a benchmark combining \emph{EEGDenoiseNet} and the \emph{DEAP} dataset \cite{koelstra2011deap}. Notice that the inputs of the BiLSTM are the time-frequency features extracted from the augmented data. 
Instead, \textit{Yin et al.} \cite{yin2022frequency} propose a cross-domain framework integrating time and frequency domain information, demonstrating that the extracted features are able to improve the performance of state-of-the-art methods when provided as input to DL models.

\section{Dataset}\label{sec:dataset}
The \emph{EEGdenoiseNet} \cite{zhang2021eegdenoisenet} is used in this study, having that it has been provided to the research community as a benchmark dataset to train and test DL-based denoising models. 

In fact, \textit{Zhang et al.} construct a dataset exploiting EEG, EOG and EMG signals of publicly available datasets, processing these data to obtain neural and noise signals that could be considered unaffected by other sources and thus \emph{clean}.\\
In particular, the authors consider \textit{Cho et al.}'s EEG dataset \cite{cho2017eeg}, presenting signals collected with 64 electrodes on 52 subjects during a motor execution and imagery experiment. \\
The EOG signals have been instead taken from \textit{Kanoga et al.} \cite{kanoga2016assessing}, and the \textit{BCI Competition IV dataset 2a and 2b} \cite{tangermann2012review}, while the EMG signals are related to a facial EMG dataset \cite{rantanen2016survey}.

Afterwards, the data are pre-processed as follows:
\begin{enumerate}
    \item Signals are notch (50 Hz) and bandpass (EEG: 1-80 Hz, EOG: 0.3-10 Hz, and EMG: 1-120 Hz) filtered.
    \item Signals are re-sampled, considering a sampling rate of 256 Hz or 512 Hz for the EEG, 256 Hz for the EOG, and 512 Hz for the EMG signals.
    \item Signals are divided in segments of 2 s to provide data as cleaner as possible, and standardized. Segments are visually inspected by experts.
\end{enumerate}
Notice that between point 1 and 2, EEG signals are processed with the independent component analysis based \emph{ICLabel} toolbox \cite{pion2019iclabel} to obtain clean ground truth data.
The data resulting from this process are 4,514, 3,400, and 5,598 \emph{pure} (as defined by the authors) EEG, EOG, and EMG segments, respectively.\\
The pure EEG data are used as the \emph{ground truth} and semi-synthetic data produced by linearly combining these data with EOG or EMG segments, according to the following formula:
\begin{equation}
    y = x + \lambda n
    \label{eq:linear_noise_signal}
\end{equation}
where $y$ is the noisy signal, $x$ the pure EEG signal, $n$ the EOG or EMG noise, and $\lambda$ a hyperparameter controlling the SNR.
For further information, please consult the original publication by \textit{Zhang et al.} \cite{zhang2021eegdenoisenet}.

\section{Proposed model}\label{sec:proposed_model}

In this work a novel denoising model leveraging data in the frequency domain is proposed.

The idea is that given a prior knowledge about the noise spectral features, an optimal convolutional filter, or a cascade of filters, can be computed to separate the noise from signal.
Therefore the proposed model is trained to learn an empirical relationship which connects the spectral characteristics of noise and noisy signal to a non-linear transformation able to denoise that signal.
The assumptions under which the model can operate are:
\begin{itemize}
	\item the PSD estimate of the noise is given;
	\item the relation between signal and noise is known. 
\end{itemize}
For the \textit{EEGdenoiseNet} dataset this relationship is a linear mixture of the clean signal and the artifacts (OAs and MAs considered one at a time) as per Eq. \ref{eq:linear_noise_signal}.

\begin{figure}[h]
    \centering
    \includegraphics[width=0.44\textwidth]{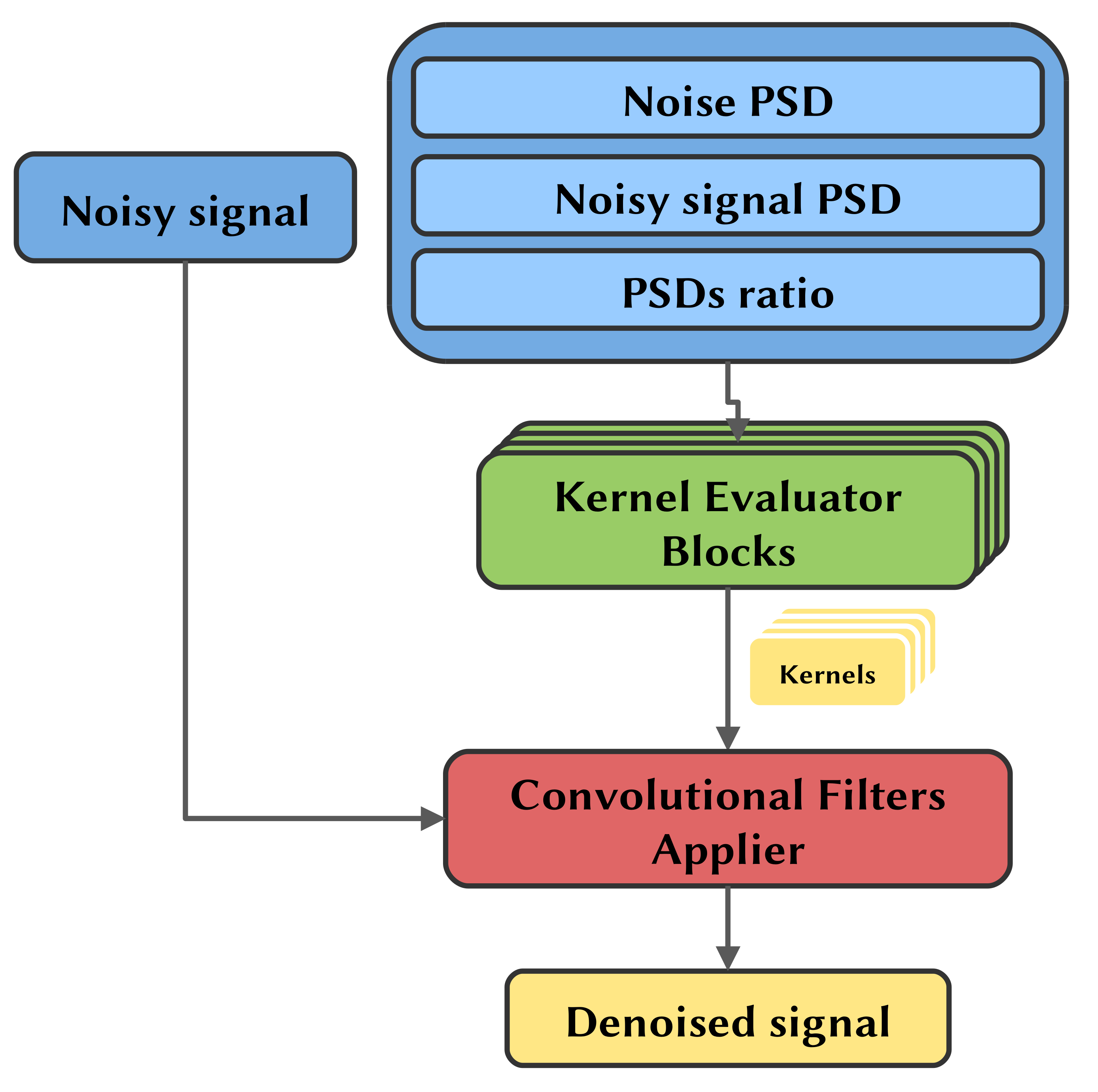}
    \caption{Model pipeline presenting the input form, the model mainly constituted by the kernel evaluator and the convolutional filter applier, and the obtained output.}
    \label{fig:model_pipeline}
\end{figure}

The PSD related to noise $\lambda n$, the PSD related to noisy signal $y$, and the noisy signal are all given separately as multiple inputs to the model, which consists of two major components, repeatedly applied: the kernel evaluator and the convolutional filters applier.
The kernel evaluator is used to evaluate the best convolutional filters from the frequencies that are characteristic of the noisy signal and noise.
The convolutional filters applier then effectively apply the filters estimated by the kernel evaluator to the time domain signal.
The overall pipeline of the model is depicted in Figure \ref{fig:model_pipeline}.

\subsection{Model inputs}
The model is inputted with two inputs not directly interacting with each other: the PSDs and the time series.
The former is the concatenation of the PSD of the pure noise, the PSD of the noisy signal (which is always known) and the ratio between the noisy signal and the pure noise PSDs, which does not add any further information but facilitates model learning since the ratio is an operation not easily reproducible by the following convolutional operations. 
The PSDs and their ratio are processed by the kernel evaluator blocks only.
The latter, i.e., the time series, is the noisy signal in the time domain, which is processed in cascade by the convolutional filters applier block in order to obtain the denoised signal.

\subsection{Kernel evaluator}
\begin{figure}[b!]
    \centering
    \includegraphics[width=0.61\textwidth]{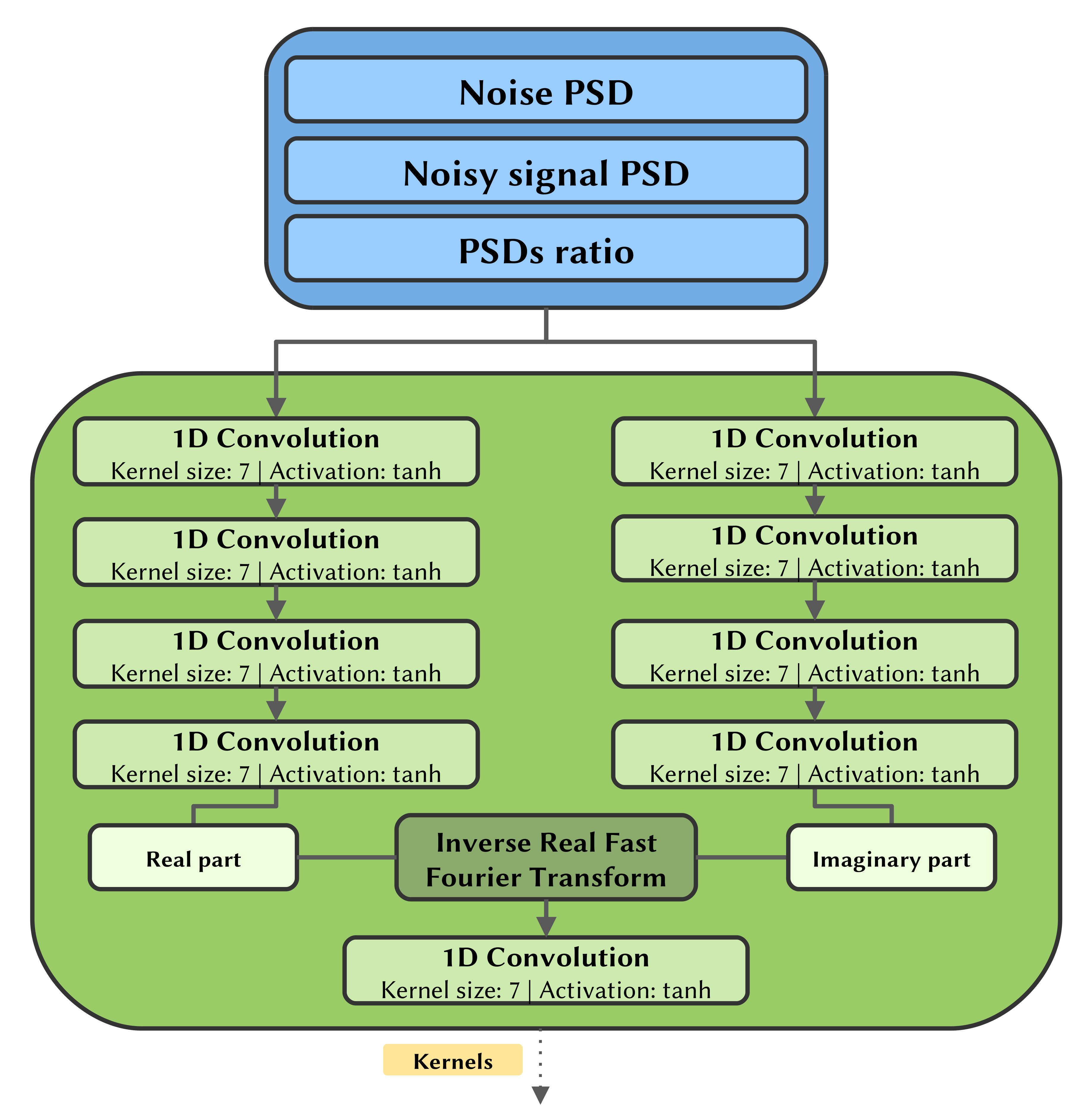}
    \caption{Kernel evaluator block diagram depicting the input form, the kernel evaluator, and its output.}
    \label{fig:kernel_evaluator}
\end{figure}

For each convolutional step a kernel evaluator block, which structure is depicted in Figure \ref{fig:kernel_evaluator}, evaluates a set of convolutional filters (i.e., the kernels values) to be applied to the time series.

Each block is inputted with the PSDs, which are processed by two symmetric series of 1D convolutional layers with \textit{tanh} activations. These branches independently estimate the real and imaginary part of the filters that will be subsequently applied to the time series. Remind that in this case the model is working in the frequency domain, dealing with PSDs.

To translate the filters into the time domain, where they will actually operate on the noisy signal, an inverse real fast Fourier transformation (IRFFT) is then applied to the complex 1D arrays obtained by assembling the two branches.

The IRFFT operation adds algorithmic capabilities to the model since the convolutional and activation layers in the pipeline cannot replace it by performing an analogous transformation.

The inverse fast Fourier transform (IFFT) is an algorithm that efficiently computes the inverse discrete Fourier transform (IDFT) of a sequence and is given as follows:

\begin{equation} \label{eq:idft_equation}
x[n] = \frac{1}{N} \sum_{k=0}^{N-1} X[k] e^{i \frac{2\pi}{N}kn}
\end{equation}

The IRFFT is a particular case of IFFT which returns real-valued sequences, as would be desired for the time-domain filters, without sacrificing the useful algorithmic capabilities of the Fourier transform.

As a final step, the filters outputted by the IRFFT are linearly combined with a 1D convolutional layer with kernel size equal to 1. The last \textit{tanh} activation forces the filters values in the range $[-1; 1]$, avoiding numerical problems coming from too high values and acting as a normalizer on filters.

The length of the filters evaluated by the kernel evaluator block is equal to the length of the input noisy signal itself, thus the filters are able to act on any signal frequency and to extract both local and global features.
The kernel evaluator blocks contain all and only the trainable parameters of the model, specifically the kernels of the 1D convolutions applied to the PSDs.
Therefore these parameters, which amount to 224192, depend only on the frequency characteristics of the signal and noise.
The choice of the number of convolutional layers and the size of the kernels they apply are dictated by having a sufficiently large receptive field that operates on correlated frequencies that are close to each other, without taking into account very distant frequencies.

\subsection{Convolutional filters applier}
\begin{figure}[h]
    \centering
    \includegraphics[width=0.70\textwidth]{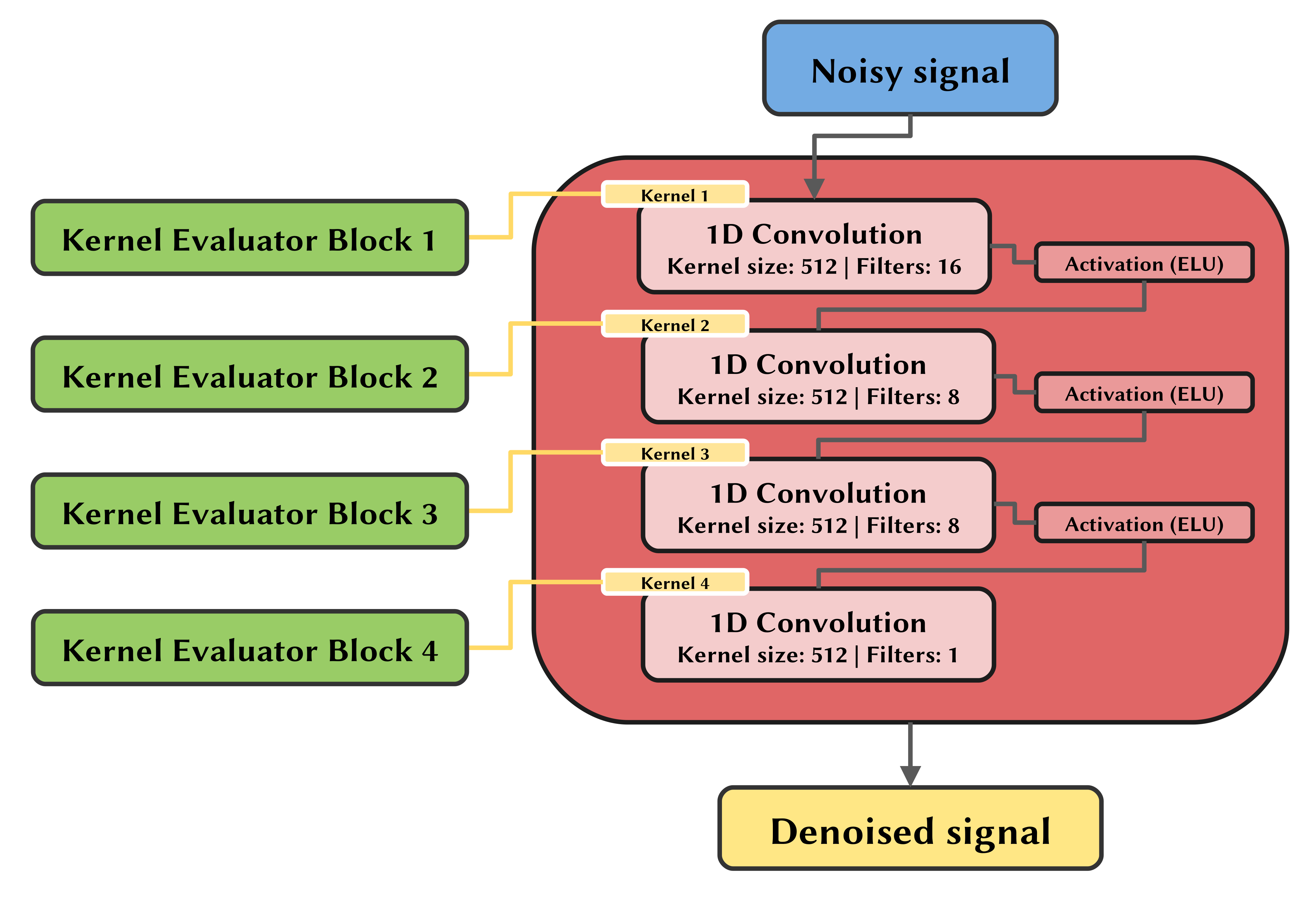}
    \caption{Convolutional filter applier schema using the previously obtained filters as inputs. The output is the denoised signal.}
    \label{fig:conv_filters_applier_schema}
\end{figure}

The filters evaluated by the kernel evaluator blocks are inputted to the convolutional filter applier, which uses them without further changes.
As depicted in Figure \ref{fig:conv_filters_applier_schema}, on the first step this part of the model applies 1D convolutions directly to the noisy signal using the filters of the first kernel evaluator block and the resulting features are inputted to an ELU activation function.
Subsequently, these features are convoluted with the filters evaluated by the second block and once again activated by the ELU function.
This operation is repeated in cascade until the last convolution, which returns directly the denoised signal. No activation is applied in this last step and the range of possible values is therefore $(-\infty; +\infty)$, which is also the range of the possible clean signals.

The number of convolutions applied in series and the number of filters applied at each convolutional step were sized based on domain knowledge and in a way that limited the total number of parameters trained in the kernel evaluator blocks.

\section{Performed experiments}\label{sec:results_discussion}

The denoising capabilities of the proposed model are evaluated on the basis of the reconstruction performance of pure EEG signals from the \textit{EEGdenoiseNet} dataset when contaminated by EOG or EMG artifacts, separately, following the paradigm proposed by the authors of the benchmark model \cite{zhang2021eegdenoisenet} and considered by most of the reported literature works (Section \ref{sec:related_works}).
In this section the data preparation method, the metrics used to evaluate the model and the training process are described.
Finally, statistical and graphical results of the model are reported.

\subsection{Data preparation}
The original dataset has been randomly partitioned into two mutually exclusive subsets, i.e., a training dataset (60$\%$) and a test dataset (40$\%$).
Therefore, the training dataset consists of 2,708 EEG, 3,358 EMG and 2,040 EOG samples while the test dataset consists of 1,806 EEG, 2,240 EMG and 1,360 EOG samples.

To synthesize the noisy signals from the pure samples, the linear relation defined in Eq. \ref{eq:linear_noise_signal} has been used. The $\lambda$ values are  randomly sampled to obtain a uniform distribution of the signal-to-noise ratio (SNR) in the range $[-7 dB; 4 dB]$ for the model training and in the range $[-7 dB; 2 dB]$ for the model testing.
This is a common range for OAs and MAs and the same range has been used by the \textit{EEGdenoiseNet} authors \cite{zhang2021eegdenoisenet}.

During the training phase, the noisy data synthesis is performed runtime in a random way, allowing the model to be trained with constantly new combinations of signal, noise and SNR.

Both noisy signal and pure noise are standardized using the mean and standard deviation of the noisy signal.

\subsection{Metrics}
In order to quantitatively evaluate the performance of the model in the denoising task, standard metrics used for benchmarking on \emph{EEGdenoiseNet} data are adopted.

The {\em Root Mean Square Error} (RMSE) is used to measure the variance between
the output predicted by the model and the ground truth and it is defined by:
\begin{equation} \label{eq:rmse}
RMSE = \sqrt{\frac{\sum_{i=1}^{N}(x_i - \tilde{y}_i)^2}{N}}
\end{equation}

\noindent where $x$ denotes the EEG signal, $\tilde{y}$ the denoised signal and $N$ the total number of data points of the signal.

To avoid a metric depending on the absolute value of the signals, the {\em Relative Root Mean Square Error} (RRMSE) is used, which in the time domain is expressed as follows:

\begin{equation} \label{eq:rrmse_temporal}
RRMSE_{t} = \sqrt{\frac{\sum_{i=1}^{N}(x_i - \tilde{y}_i)^2}{\sum_{i=1}^{N} x_i^2}}
\end{equation}

\noindent while when considering the frequency domain we obtain the following:

\begin{equation} \label{eq:rrmse_spectral}
RRMSE_{f} = \sqrt{\frac{\sum_{i=1}^{N}(PSD(x)_i - PSD(\tilde{y})_i)^2}{\sum_{i=1}^{N} PSD(x)_i^2}}
\end{equation}

The {\em correlation coefficient} (CC), also referred to as {\em Pearson correlation coefficient}, measures the degree of the statistical relationship between two variables, in this case the ground truth signal and the denoised signal. The CC takes values in the range $[-1; 1]$, where $\pm1$ indicates complete linear dependence between the variables, while  $0$ could mean their independence, and is defined as follows:

\begin{equation} \label{CC}
CC = \frac{\sum_{i=1}^{n}(x_i - \overline{x})(\tilde{y}_i - \overline{y})}{\sqrt{\sum_{i=1}^{n}(x_i - \overline{x})^2}\sqrt{\sum_{i=1}^{n}(\tilde{y}_i - \overline{y})^2}}
\end{equation}

\noindent where $\overline{x}$ and $\overline{y}$ are the means of the ground truth and of the model output, respectively.

\subsection{Training process}

Standardization helps to speed up the training process since input centering and scaling operations improve the rate at which the neural network converges \cite{lecun2002efficient}. Indeed, the learning algorithm is sensitive to the input scale, and if the input data are not standardized, it may take longer for the algorithm to find a good set of parameters, \textit{i.e.}, weights and thresholds of the network, or the learning algorithm may get stuck in a local minima.
Moreover, the standardization of the input data makes the model capable of processing EEG signals with wider amplitude ranges.
Nevertheless, only the mean and standard deviation of the noisy signal are always known. Therefore, the noisy signal $y$, the pure EEG signal $x$ and the pure noise $n$ are processed in a similar manner according to:

\begin{equation} \label{eq:input_standardization}
\hat{x} = \frac{x - \overline{y}}{\sigma_y},
\quad
\hat{y} = \frac{y - \overline{y}}{\sigma_y},
\quad
\hat{n} = \frac{n - \overline{y}}{\sigma_y}
\end{equation}

These signals, as well as the PSDs evaluated by them, are the actual inputs of the model.

The loss function minimized in the training phase is a combination of three different terms, differently weighted:

\begin{equation} \label{eq:loss}
L = a L_{RRMSE_{t}} + b L_{RRMSE_{f}} + c L_{log-cosh}
\end{equation}

\noindent where $L_{RRMSE_{t}}$ is the temporal RRMSE defined in Eq. \ref{eq:rrmse_temporal}, $L_{RRMSE_{f}}$ is the spectral RRMSE defined in Eq. \ref{eq:rrmse_spectral}, and $L_{log-cosh}$ the log-cosh error of the ground truth and predicted signals in the time domain, which is defined by the equation:

\begin{equation}
L_{log-cosh}(x, \tilde{y}) = \sum_{i=1}^{N}\log(\cosh(\tilde{y}_i - x_i))
\end{equation}

This function provides a smooth approximation to the mean absolute error for values near 0.
The $a$, $b$, and $c$ coefficients are empirically chosen for the training and are equal to $0.25$, $0.25$, and $0.5$, respectively.

The optimization method used to find the best weights of the kernel evaluator blocks is AdaMax \cite{kingma2014adam}.

The code was developed in Python 3.8.10 and the proposed model was designed using the TensorFlow library, version 2.8.
The experiments were run on an Nvidia Quadro RTX 4000 for a total training time of 61 hours.

\subsection{Results and discussions}

A single model has been trained on both EMG and EOG artifacts at the same time in order to have a solution capable of handling both cases. In fact, the noisy signals are affected by either OAs or MAs. Therefore, both EMG only and EOG only affected signals are inputted to the model, as introduced at the beginning of Section \ref{sec:results_discussion}.

\begin{figure}[b]
    \centering
    \includegraphics[width=0.68\textwidth]{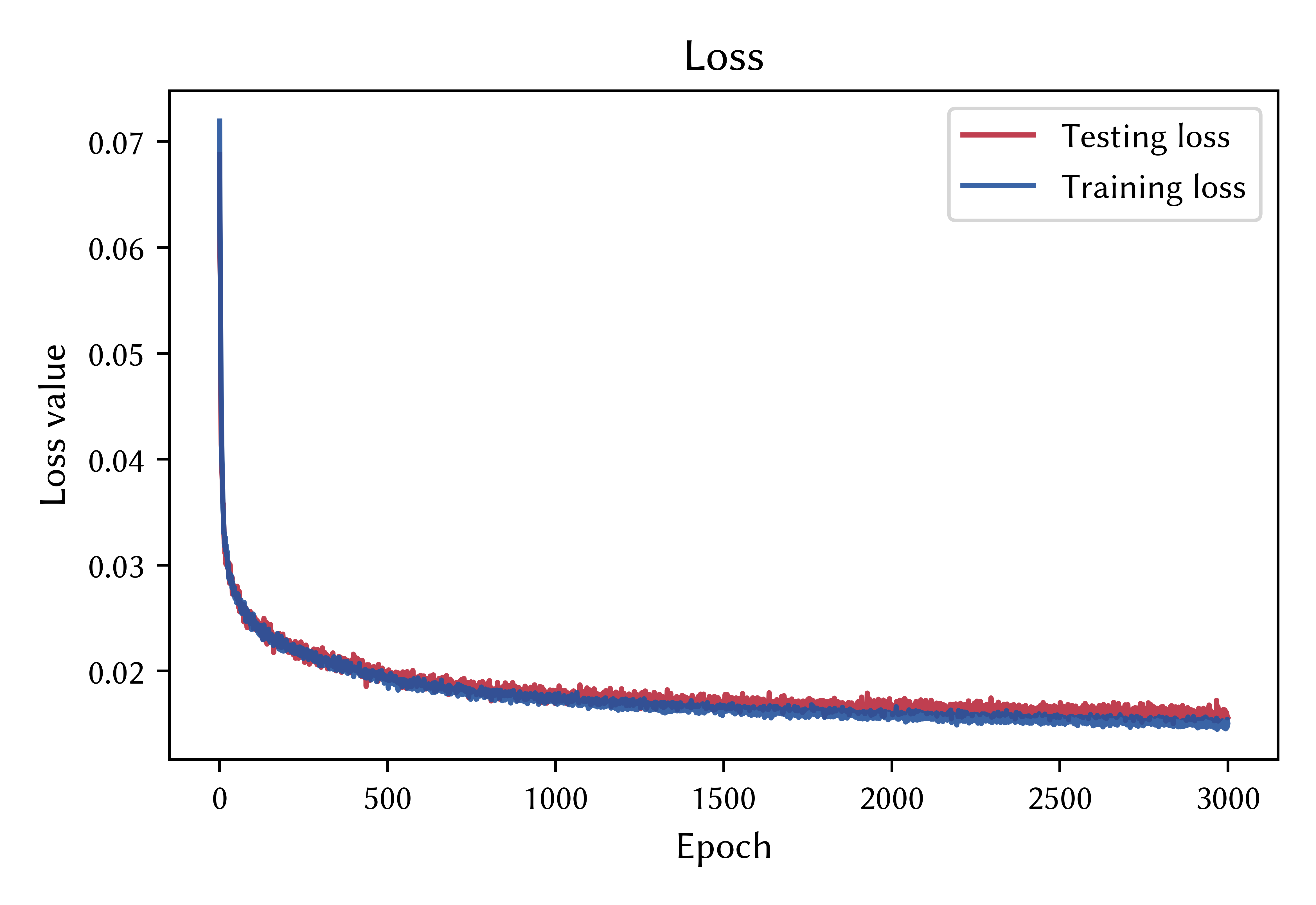}
    \caption{Loss of the training set and test set as training epochs increase.}
    \label{fig:loss_trend}
\end{figure}

Figure \ref{fig:loss_trend} shows the trend of loss as the epochs change for the training and test sets.
For both datasets the loss values monotonically decrease and no significant overfit is present, indication that the model design and the runtime data synthesis approach used are effective in avoiding this issue.

\begin{figure}[h]
    \centering
    \includegraphics[width=1.0\textwidth]{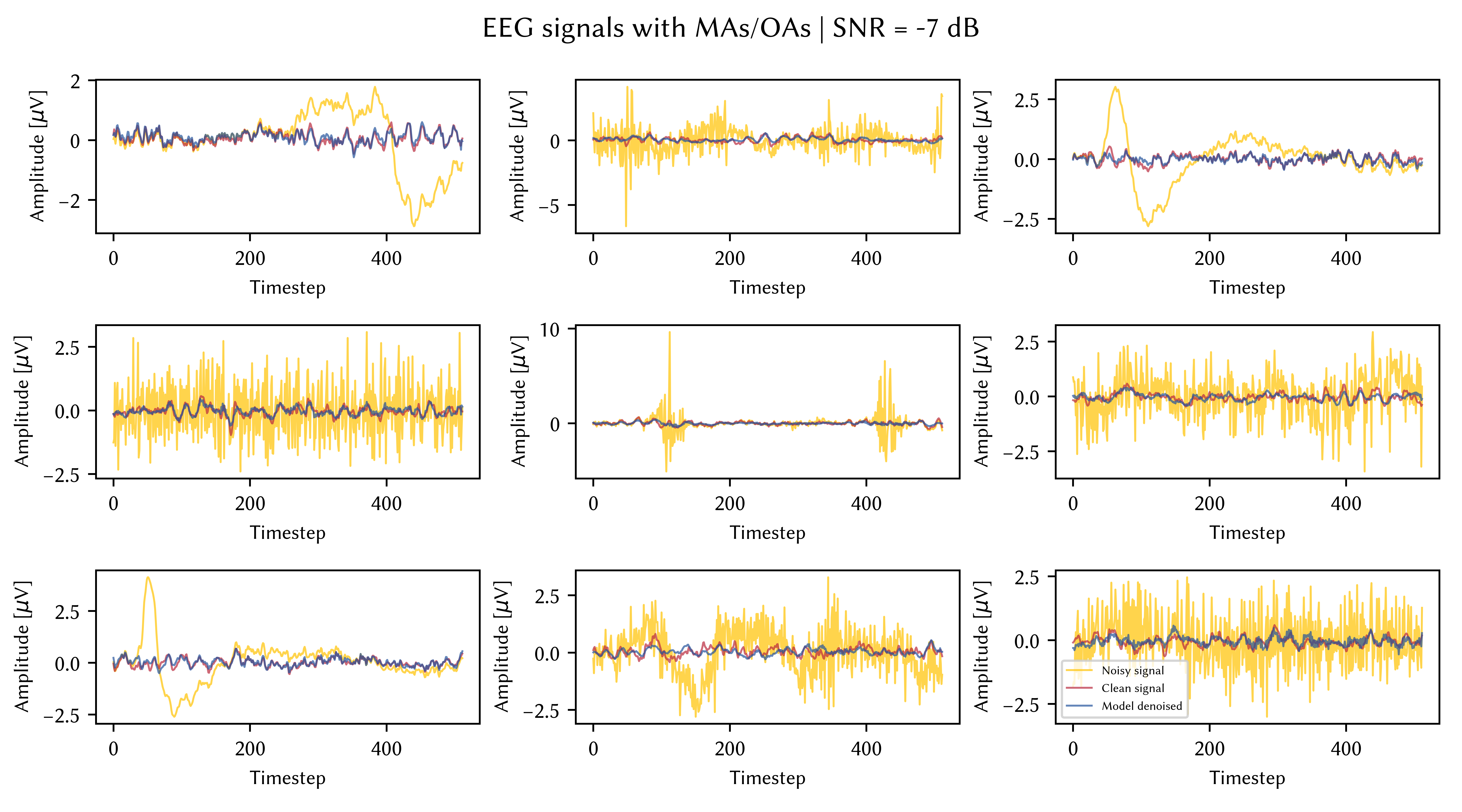}
    \caption{Test examples of noisy EEG signals corrupted by muscular or ocular artifacts with a signal-to-noise ratio equal to -7 $dB$ denoised by the model. The timestep denotes the signal sample.}
    \label{fig:eog_emg_signal_denoised}
\end{figure}

Using the last epoch weights, we qualitatively demonstrate the denoising capabilities of the model on EOG and EMG artifacts in Figure \ref{fig:eog_emg_signal_denoised}.
We can observe that both the high frequencies related to MAs and low frequencies related to OAs are filtered by the model while the majority of the structures related to the true EEG signals are preserved. Moreover, several amplitude ranges are properly managed by the model as well as different SNRs.

The quantitative results of the proposed model and of the literature benchmark models for MAs and OAs are reported in Table \ref{tab:model_results_emg} and \ref{tab:model_results_eog}, respectively.

%

\begin{table}
\centering
\begin{tabular}{lccc} 
	\multicolumn{4}{c}{\textbf{Muscular artifacts}}\\ \toprule
    {Model} & {$RRMSE_{t}$} & {$RRMSE_{f}$} & {$CC$} \\ \midrule
    FCNN \cite{zhang2021novel} & 0.585  & 0.580 & 0.796 \\
    Simple CNN \cite{zhang2021novel} & 0.646  & 0.649 & 0.783 \\
    Complex CNN \cite{zhang2021novel} & 0.650  & 0.633 & 0.780 \\
    RNN \cite{zhang2021novel} & 0.570  & 0.530 & 0.812 \\
    Novel CNN \cite{zhang2021novel} & 0.448  & 0.442 & 0.863 \\
    
    DeepSeparator \cite{yu2022embedding} & 0.712  & 0.717 & 0.734 \\
    
    EEGDnet \cite{pu2022eegdnet} & 0.677  & 0.626 & 0.732 \\

    Proposed model  & 0.573 & 0.496 & 0.805 \\
    
    \bottomrule
\end{tabular}
	\caption{Metrics of the proposed denoising model and benchmark models averaged on all SNRs ($[-7 dB; 2 dB]$) with noisy signals affected by MAs.}
	\label{tab:model_results_emg}
\end{table}

\begin{table}
\centering
\begin{tabular}{lccc} 
	\multicolumn{4}{c}{\textbf{Ocular artifacts}}\\ \toprule
    {Model} & {$RRMSE_{t}$} & {$RRMSE_{f}$} & {$CC$} \\ \midrule
    DeepSeparator \cite{yu2022embedding} & 0.705  & 0.747 & 0.769 \\
    
    EEGDnet \cite{pu2022eegdnet} & 0.497  & 0.491 & 0.868 \\

    Proposed model  & 0.405  & 0.490 & 0.917 \\
    \bottomrule
\end{tabular}
	\caption{Metrics of the proposed denoising model and benchmark models averaged on all SNRs ($[-7 dB; 2 dB]$) with noisy signals affected by OAs.}
	\label{tab:model_results_eog}
\end{table}

Regarding EEG affected by MAs, the reported values are in line with the benchmark models results.
The proposed model demonstrated robust performance, achieving the third best results in the temporal domain metrics, $RRMSE_{t}$ and $CC$, and the second best result in the spectral metric $RRMSE_{f}$. This latter metric provides insight into the method ability to effectively preserve the spectral information of the clean EEG signal.

Concerning OAs, our method achieves the best performance on all three metrics among the reported benchmark models.
Furthermore, the similarity of the proposed model results in $RRMSE_{f}$ values on both muscular and ocular artifacts highlights the method stability and its ability to perform indiscriminately well on both high and low frequencies.

In contrast to the current state of the art, the proposed model is able to achieve these results using not only a single architecture but also a single set of trained parameters, valid on both MAs and OAs. This allows the two main sources of noise in EEG signals to be handled with a single solution.
However, while the need to have as input an estimate of the PSD of the noise afflicting the signal makes the method more adaptable to new noise conditions with known characteristics, it also makes it sensitive to precise \textit{a priori} knowledge of the frequency characteristics of this noise that cannot be easily estimated.
In the future we will develop methods trained on the reference dataset to best estimate the PSD of the noise of each signal as a substitute for the exact PSD provided in the current work.
Moreover, different activation functions and artifact amount estimation methodologies \cite{vialatte2008eeg} will be considered to provide a better denoising of the EEG signals.

\section{Conclusions}\label{sec:conclusions}
The significant issue of muscle and ocular artifact removal in EEG data has been tackled in this research.
We introduced a unique solution capable of dealing with both artifact types using a single model.
Our proposed method leverages dynamically assessed convolutional filters, which are determined based on the frequency features of the noise and the noisy signal.
With this knowledge, our model has proven its effectiveness in both qualitatively and quantitatively cleaning EEG signals from muscle and ocular artifacts.
This achievement either matches or exceeds the performance of existing state-of-the-art models, which typically require specific training on either ocular or muscle artifacts.
Thus, this study marks a substantial step forward in EEG data processing by offering a versatile spectral-based strategy for artifacts elimination and provides a baseline for subsequent work addressing the problem of estimating noise frequencies, whether with experimental solutions, algorithmic solutions, or a combination of the two.

\begin{acknowledgments}
\noindent  This work was partially supported by the MUR under the grant “Dipartimenti di Eccellenza 2023-2027" of the Department of Informatics, Systems and Communication of the University of Milano-Bicocca, Italy.

\end{acknowledgments}

\bibliography{mybib}

\begin{thebibliography}{29}
\expandafter\ifx\csname natexlab\endcsname\relax\def\natexlab#1{#1}\fi
\providecommand{\url}[1]{\texttt{#1}}
\providecommand{\href}[2]{#2}
\providecommand{\path}[1]{#1}
\providecommand{\DOIprefix}{doi:}
\providecommand{\ArXivprefix}{arXiv:}
\providecommand{\URLprefix}{URL: }
\providecommand{\Pubmedprefix}{pmid:}
\providecommand{\doi}[1]{\href{http://dx.doi.org/#1}{\path{#1}}}
\providecommand{\Pubmed}[1]{\href{pmid:#1}{\path{#1}}}
\providecommand{\bibinfo}[2]{#2}
\ifx\xfnm\relax \def\xfnm[#1]{\unskip,\space#1}\fi
\bibitem[{Saibene et~al.(2023)Saibene, Caglioni, Corchs, and
  Gasparini}]{saibene2023eeg}
\bibinfo{author}{A.~Saibene}, \bibinfo{author}{M.~Caglioni},
  \bibinfo{author}{S.~Corchs}, \bibinfo{author}{F.~Gasparini},
\newblock \bibinfo{title}{Eeg-based bcis on motor imagery paradigm using
  wearable technologies: A systematic review},
\newblock \bibinfo{journal}{Sensors} \bibinfo{volume}{23}
  (\bibinfo{year}{2023}) \bibinfo{pages}{2798}.
\bibitem[{Zhang et~al.(2020)Zhang, Yin, Chen, and Nichele}]{zhang2020emotion}
\bibinfo{author}{J.~Zhang}, \bibinfo{author}{Z.~Yin},
  \bibinfo{author}{P.~Chen}, \bibinfo{author}{S.~Nichele},
\newblock \bibinfo{title}{Emotion recognition using multi-modal data and
  machine learning techniques: A tutorial and review},
\newblock \bibinfo{journal}{Information Fusion} \bibinfo{volume}{59}
  (\bibinfo{year}{2020}) \bibinfo{pages}{103--126}.
\bibitem[{Urig{\"u}en and Garcia-Zapirain(2015)}]{uriguen2015eeg}
\bibinfo{author}{J.~A. Urig{\"u}en}, \bibinfo{author}{B.~Garcia-Zapirain},
\newblock \bibinfo{title}{Eeg artifact removal—state-of-the-art and
  guidelines},
\newblock \bibinfo{journal}{Journal of neural engineering} \bibinfo{volume}{12}
  (\bibinfo{year}{2015}) \bibinfo{pages}{031001}.
\bibitem[{Rashmi and Shantala(2022)}]{rashmi2022eeg}
\bibinfo{author}{C.~Rashmi}, \bibinfo{author}{C.~Shantala},
\newblock \bibinfo{title}{Eeg artifacts detection and removal techniques for
  brain computer interface applications: a systematic review},
\newblock \bibinfo{journal}{International Journal of Advanced Technology and
  Engineering Exploration} \bibinfo{volume}{9} (\bibinfo{year}{2022})
  \bibinfo{pages}{354}.
\bibitem[{Jiang et~al.(2019)Jiang, Bian, and Tian}]{jiang2019removal}
\bibinfo{author}{X.~Jiang}, \bibinfo{author}{G.-B. Bian},
  \bibinfo{author}{Z.~Tian},
\newblock \bibinfo{title}{Removal of artifacts from eeg signals: a review},
\newblock \bibinfo{journal}{Sensors} \bibinfo{volume}{19}
  (\bibinfo{year}{2019}) \bibinfo{pages}{987}.
\bibitem[{Mumtaz et~al.(2021)Mumtaz, Rasheed, and Irfan}]{mumtaz2021review}
\bibinfo{author}{W.~Mumtaz}, \bibinfo{author}{S.~Rasheed},
  \bibinfo{author}{A.~Irfan},
\newblock \bibinfo{title}{Review of challenges associated with the eeg artifact
  removal methods},
\newblock \bibinfo{journal}{Biomedical Signal Processing and Control}
  \bibinfo{volume}{68} (\bibinfo{year}{2021}) \bibinfo{pages}{102741}.
\bibitem[{Wallstrom et~al.(2004)Wallstrom, Kass, Miller, Cohn, and
  Fox}]{wallstrom2004automatic}
\bibinfo{author}{G.~L. Wallstrom}, \bibinfo{author}{R.~E. Kass},
  \bibinfo{author}{A.~Miller}, \bibinfo{author}{J.~F. Cohn},
  \bibinfo{author}{N.~A. Fox},
\newblock \bibinfo{title}{Automatic correction of ocular artifacts in the eeg:
  a comparison of regression-based and component-based methods},
\newblock \bibinfo{journal}{International journal of psychophysiology}
  \bibinfo{volume}{53} (\bibinfo{year}{2004}) \bibinfo{pages}{105--119}.
\bibitem[{Mannan et~al.(2018)Mannan, Kamran, and
  Jeong}]{mannan2018identification}
\bibinfo{author}{M.~M.~N. Mannan}, \bibinfo{author}{M.~A. Kamran},
  \bibinfo{author}{M.~Y. Jeong},
\newblock \bibinfo{title}{Identification and removal of physiological artifacts
  from electroencephalogram signals: A review},
\newblock \bibinfo{journal}{Ieee Access} \bibinfo{volume}{6}
  (\bibinfo{year}{2018}) \bibinfo{pages}{30630--30652}.
\bibitem[{Chen et~al.(2019)Chen, Xu, Liu, Lee, Chen, Zhang, McKeown, and
  Wang}]{chen2019removal}
\bibinfo{author}{X.~Chen}, \bibinfo{author}{X.~Xu}, \bibinfo{author}{A.~Liu},
  \bibinfo{author}{S.~Lee}, \bibinfo{author}{X.~Chen},
  \bibinfo{author}{X.~Zhang}, \bibinfo{author}{M.~J. McKeown},
  \bibinfo{author}{Z.~J. Wang},
\newblock \bibinfo{title}{Removal of muscle artifacts from the eeg: A review
  and recommendations},
\newblock \bibinfo{journal}{IEEE Sensors Journal} \bibinfo{volume}{19}
  (\bibinfo{year}{2019}) \bibinfo{pages}{5353--5368}.
\bibitem[{Zhang et~al.(2021)Zhang, Zhao, Wei, Mantini, Li, and
  Liu}]{zhang2021eegdenoisenet}
\bibinfo{author}{H.~Zhang}, \bibinfo{author}{M.~Zhao},
  \bibinfo{author}{C.~Wei}, \bibinfo{author}{D.~Mantini},
  \bibinfo{author}{Z.~Li}, \bibinfo{author}{Q.~Liu},
\newblock \bibinfo{title}{Eegdenoisenet: a benchmark dataset for deep learning
  solutions of eeg denoising},
\newblock \bibinfo{journal}{Journal of Neural Engineering} \bibinfo{volume}{18}
  (\bibinfo{year}{2021}) \bibinfo{pages}{056057}.
\bibitem[{Minguillon et~al.(2017)Minguillon, Lopez-Gordo, and
  Pelayo}]{minguillon2017trends}
\bibinfo{author}{J.~Minguillon}, \bibinfo{author}{M.~A. Lopez-Gordo},
  \bibinfo{author}{F.~Pelayo},
\newblock \bibinfo{title}{Trends in eeg-bci for daily-life: Requirements for
  artifact removal},
\newblock \bibinfo{journal}{Biomedical Signal Processing and Control}
  \bibinfo{volume}{31} (\bibinfo{year}{2017}) \bibinfo{pages}{407--418}.
\bibitem[{Zhang et~al.(2021)Zhang, Wei, Zhao, Liu, and Wu}]{zhang2021novel}
\bibinfo{author}{H.~Zhang}, \bibinfo{author}{C.~Wei},
  \bibinfo{author}{M.~Zhao}, \bibinfo{author}{Q.~Liu}, \bibinfo{author}{H.~Wu},
\newblock \bibinfo{title}{A novel convolutional neural network model to remove
  muscle artifacts from eeg},
\newblock in: \bibinfo{booktitle}{ICASSP 2021-2021 IEEE International
  Conference on Acoustics, Speech and Signal Processing (ICASSP)},
  \bibinfo{organization}{IEEE}, \bibinfo{year}{2021}, pp.
  \bibinfo{pages}{1265--1269}.
\bibitem[{Yu et~al.(2022)Yu, Li, Lou, Wei, and Liu}]{yu2022embedding}
\bibinfo{author}{J.~Yu}, \bibinfo{author}{C.~Li}, \bibinfo{author}{K.~Lou},
  \bibinfo{author}{C.~Wei}, \bibinfo{author}{Q.~Liu},
\newblock \bibinfo{title}{Embedding decomposition for artifacts removal in eeg
  signals},
\newblock \bibinfo{journal}{Journal of Neural Engineering} \bibinfo{volume}{19}
  (\bibinfo{year}{2022}) \bibinfo{pages}{026052}.
\bibitem[{Pu et~al.(2022)Pu, Yi, Chen, Ma, Zhao, and Ren}]{pu2022eegdnet}
\bibinfo{author}{X.~Pu}, \bibinfo{author}{P.~Yi}, \bibinfo{author}{K.~Chen},
  \bibinfo{author}{Z.~Ma}, \bibinfo{author}{D.~Zhao}, \bibinfo{author}{Y.~Ren},
\newblock \bibinfo{title}{Eegdnet: Fusing non-local and local self-similarity
  for eeg signal denoising with transformer},
\newblock \bibinfo{journal}{Computers in Biology and Medicine}
  \bibinfo{volume}{151} (\bibinfo{year}{2022}) \bibinfo{pages}{106248}.
\bibitem[{Wang et~al.(2022)Wang, Li, and Wang}]{wang2022novel}
\bibinfo{author}{W.~Wang}, \bibinfo{author}{B.~Li}, \bibinfo{author}{H.~Wang},
\newblock \bibinfo{title}{A novel end-to-end network based on a bidirectional
  gru and a self-attention mechanism for denoising of electroencephalography
  signals},
\newblock \bibinfo{journal}{Neuroscience} \bibinfo{volume}{505}
  (\bibinfo{year}{2022}) \bibinfo{pages}{10--20}.
\bibitem[{Zhang et~al.(2022)Zhang, Yu, Rong, and Iwata}]{zhang2022novel}
\bibinfo{author}{Z.~Zhang}, \bibinfo{author}{X.~Yu}, \bibinfo{author}{X.~Rong},
  \bibinfo{author}{M.~Iwata},
\newblock \bibinfo{title}{A novel multimodule neural network for eeg
  denoising},
\newblock \bibinfo{journal}{IEEE Access} \bibinfo{volume}{10}
  (\bibinfo{year}{2022}) \bibinfo{pages}{49528--49541}.
\bibitem[{Brophy et~al.(2022)Brophy, Redmond, Fleury, De~Vos, Boylan, and
  Ward}]{brophy2022denoising}
\bibinfo{author}{E.~Brophy}, \bibinfo{author}{P.~Redmond},
  \bibinfo{author}{A.~Fleury}, \bibinfo{author}{M.~De~Vos},
  \bibinfo{author}{G.~Boylan}, \bibinfo{author}{T.~Ward},
\newblock \bibinfo{title}{Denoising eeg signals for real-world bci applications
  using gans},
\newblock \bibinfo{journal}{Frontiers in Neuroergonomics} \bibinfo{volume}{2}
  (\bibinfo{year}{2022}) \bibinfo{pages}{44}.
\bibitem[{Wang et~al.(2022)Wang, Luo, and Shen}]{wang2022improved}
\bibinfo{author}{S.~Wang}, \bibinfo{author}{Y.~Luo}, \bibinfo{author}{H.~Shen},
\newblock \bibinfo{title}{An improved generative adversarial network for
  denoising eeg signals of brain-computer interface systems},
\newblock in: \bibinfo{booktitle}{2022 China Automation Congress (CAC)},
  \bibinfo{organization}{IEEE}, \bibinfo{year}{2022}, pp.
  \bibinfo{pages}{6498--6502}.
\bibitem[{Ozdemir et~al.(2022)Ozdemir, Kizilisik, and
  Guren}]{ozdemir2022removal}
\bibinfo{author}{M.~A. Ozdemir}, \bibinfo{author}{S.~Kizilisik},
  \bibinfo{author}{O.~Guren},
\newblock \bibinfo{title}{Removal of ocular artifacts in eeg using deep
  learning},
\newblock in: \bibinfo{booktitle}{2022 Medical Technologies Congress
  (TIPTEKNO)}, \bibinfo{organization}{IEEE}, \bibinfo{year}{2022}, pp.
  \bibinfo{pages}{1--6}.
\bibitem[{Koelstra et~al.(2011)Koelstra, Muhl, Soleymani, Lee, Yazdani,
  Ebrahimi, Pun, Nijholt, and Patras}]{koelstra2011deap}
\bibinfo{author}{S.~Koelstra}, \bibinfo{author}{C.~Muhl},
  \bibinfo{author}{M.~Soleymani}, \bibinfo{author}{J.-S. Lee},
  \bibinfo{author}{A.~Yazdani}, \bibinfo{author}{T.~Ebrahimi},
  \bibinfo{author}{T.~Pun}, \bibinfo{author}{A.~Nijholt},
  \bibinfo{author}{I.~Patras},
\newblock \bibinfo{title}{Deap: A database for emotion analysis; using
  physiological signals},
\newblock \bibinfo{journal}{IEEE transactions on affective computing}
  \bibinfo{volume}{3} (\bibinfo{year}{2011}) \bibinfo{pages}{18--31}.
\bibitem[{Yin et~al.(2022)Yin, Liu, Li, Qian, and Chen}]{yin2022frequency}
\bibinfo{author}{J.~Yin}, \bibinfo{author}{A.~Liu}, \bibinfo{author}{C.~Li},
  \bibinfo{author}{R.~Qian}, \bibinfo{author}{X.~Chen},
\newblock \bibinfo{title}{Frequency information enhanced deep eeg denoising
  network for ocular artifact removal},
\newblock \bibinfo{journal}{IEEE Sensors Journal} \bibinfo{volume}{22}
  (\bibinfo{year}{2022}) \bibinfo{pages}{21855--21865}.
\bibitem[{Cho et~al.(2017)Cho, Ahn, Ahn, Kwon, and Jun}]{cho2017eeg}
\bibinfo{author}{H.~Cho}, \bibinfo{author}{M.~Ahn}, \bibinfo{author}{S.~Ahn},
  \bibinfo{author}{M.~Kwon}, \bibinfo{author}{S.~C. Jun},
\newblock \bibinfo{title}{Eeg datasets for motor imagery brain--computer
  interface},
\newblock \bibinfo{journal}{GigaScience} \bibinfo{volume}{6}
  (\bibinfo{year}{2017}) \bibinfo{pages}{gix034}.
\bibitem[{Kanoga et~al.(2016)Kanoga, Nakanishi, and
  Mitsukura}]{kanoga2016assessing}
\bibinfo{author}{S.~Kanoga}, \bibinfo{author}{M.~Nakanishi},
  \bibinfo{author}{Y.~Mitsukura},
\newblock \bibinfo{title}{Assessing the effects of voluntary and involuntary
  eyeblinks in independent components of electroencephalogram},
\newblock \bibinfo{journal}{Neurocomputing} \bibinfo{volume}{193}
  (\bibinfo{year}{2016}) \bibinfo{pages}{20--32}.
\bibitem[{Tangermann et~al.(2012)Tangermann, M{\"u}ller, Aertsen, Birbaumer,
  Braun, Brunner, Leeb, Mehring, Miller, Mueller-Putz
  et~al.}]{tangermann2012review}
\bibinfo{author}{M.~Tangermann}, \bibinfo{author}{K.-R. M{\"u}ller},
  \bibinfo{author}{A.~Aertsen}, \bibinfo{author}{N.~Birbaumer},
  \bibinfo{author}{C.~Braun}, \bibinfo{author}{C.~Brunner},
  \bibinfo{author}{R.~Leeb}, \bibinfo{author}{C.~Mehring},
  \bibinfo{author}{K.~J. Miller}, \bibinfo{author}{G.~Mueller-Putz}, et~al.,
\newblock \bibinfo{title}{Review of the bci competition iv},
\newblock \bibinfo{journal}{Frontiers in neuroscience}  (\bibinfo{year}{2012})
  \bibinfo{pages}{55}.
\bibitem[{Rantanen et~al.(2016)Rantanen, Ilves, Vehkaoja, Kontunen, Lylykangas,
  M{\"a}kel{\"a}, Rautiainen, Surakka, and Lekkala}]{rantanen2016survey}
\bibinfo{author}{V.~Rantanen}, \bibinfo{author}{M.~Ilves},
  \bibinfo{author}{A.~Vehkaoja}, \bibinfo{author}{A.~Kontunen},
  \bibinfo{author}{J.~Lylykangas}, \bibinfo{author}{E.~M{\"a}kel{\"a}},
  \bibinfo{author}{M.~Rautiainen}, \bibinfo{author}{V.~Surakka},
  \bibinfo{author}{J.~Lekkala},
\newblock \bibinfo{title}{A survey on the feasibility of surface emg in facial
  pacing},
\newblock in: \bibinfo{booktitle}{2016 38th Annual International Conference of
  the IEEE Engineering in Medicine and Biology Society (EMBC)},
  \bibinfo{organization}{IEEE}, \bibinfo{year}{2016}, pp.
  \bibinfo{pages}{1688--1691}.
\bibitem[{Pion-Tonachini et~al.(2019)Pion-Tonachini, Kreutz-Delgado, and
  Makeig}]{pion2019iclabel}
\bibinfo{author}{L.~Pion-Tonachini}, \bibinfo{author}{K.~Kreutz-Delgado},
  \bibinfo{author}{S.~Makeig},
\newblock \bibinfo{title}{Iclabel: An automated electroencephalographic
  independent component classifier, dataset, and website},
\newblock \bibinfo{journal}{NeuroImage} \bibinfo{volume}{198}
  (\bibinfo{year}{2019}) \bibinfo{pages}{181--197}.
\bibitem[{LeCun et~al.(2002)LeCun, Bottou, Orr, and
  M{\"u}ller}]{lecun2002efficient}
\bibinfo{author}{Y.~LeCun}, \bibinfo{author}{L.~Bottou}, \bibinfo{author}{G.~B.
  Orr}, \bibinfo{author}{K.-R. M{\"u}ller},
\newblock \bibinfo{title}{Efficient backprop},
\newblock in: \bibinfo{booktitle}{Neural networks: Tricks of the trade},
  \bibinfo{publisher}{Springer}, \bibinfo{year}{2002}, pp.
  \bibinfo{pages}{9--50}.
\bibitem[{Kingma and Ba(2014)}]{kingma2014adam}
\bibinfo{author}{D.~P. Kingma}, \bibinfo{author}{J.~Ba},
\newblock \bibinfo{title}{Adam: A method for stochastic optimization},
\newblock \bibinfo{journal}{arXiv preprint arXiv:1412.6980}
  (\bibinfo{year}{2014}).
\bibitem[{Vialatte et~al.(2008)Vialatte, Sol{\'e}-Casals, and
  Cichocki}]{vialatte2008eeg}
\bibinfo{author}{F.-B. Vialatte}, \bibinfo{author}{J.~Sol{\'e}-Casals},
  \bibinfo{author}{A.~Cichocki},
\newblock \bibinfo{title}{Eeg windowed statistical wavelet scoring for
  evaluation and discrimination of muscular artifacts},
\newblock \bibinfo{journal}{Physiological Measurement} \bibinfo{volume}{29}
  (\bibinfo{year}{2008}) \bibinfo{pages}{1435}.

\end{thebibliography}

\end{document}